\documentclass{article}

\usepackage{microtype}
\usepackage{graphicx}
\usepackage{subfigure}
\usepackage{booktabs} %
\usepackage{enumitem}
\usepackage{tcolorbox}
\usepackage{adjustbox}
\usepackage{amsmath}
\usepackage{subfigure}
\usepackage{multirow} 
\usepackage{colortbl}
\usepackage{hyperref}
\hypersetup{breaklinks=true}

\usepackage[accepted]{icml2025}

\usepackage{amsmath}
\usepackage{amssymb}
\usepackage{mathtools}
\usepackage{amsthm}

\usepackage[capitalize,noabbrev]{cleveref}

\theoremstyle{plain}

\theoremstyle{definition}

\theoremstyle{remark}

\usepackage[textsize=tiny]{todonotes}

\icmltitlerunning{Learning Distribution-Wise Control in Representation Space for Language Models}
\DeclareMathOperator*{\argmax}{arg\,max}

\begin{document}

\twocolumn[
\icmltitle{Learning Distribution-Wise Control in Representation Space \\for Language Models}

\icmlsetsymbol{equal}{*}

\begin{icmlauthorlist}
\icmlauthor{Chunyuan Deng}{yyy}
\icmlauthor{Ruidi Chang}{yyy}
\icmlauthor{Hanjie Chen}{yyy}

\end{icmlauthorlist}

\icmlaffiliation{yyy}{Department of Computer Science, Rice University}
\icmlcorrespondingauthor{Chunyuan Deng}{chunyuan.deng@rice.edu}
\icmlcorrespondingauthor{Hanjie Chen}{hanjie@rice.edu}

\icmlkeywords{Machine Learning, ICML}

\vskip 0.3in
]

\printAffiliationsAndNotice{} %

\begin{abstract}
\label{sec:abstract}
Interventions in language models (LMs) are applied strategically to steer model behavior during the forward pass. Learnable interventions, also known as representation fine-tuning, aim to apply pointwise control within the concept subspace and have proven effective in altering high-level behaviors. In this work, we extend this approach to the distribution level, enabling the model to learn not only pointwise transformations but also the surrounding regions of the concept subspace. We demonstrate that these methods perform effectively in early layers, with larger standard deviations correlating strongly with improved performance. Across eight commonsense reasoning and seven arithmetic reasoning benchmarks, our distribution-wise interventions consistently outperform pointwise interventions in controllability and robustness. These results illustrate that distribution-wise interventions provide a more comprehensive method for steering model behavior and enabling finer-grained control over language models. The code is at: \href{https://github.com/chili-lab/D-Intervention}{https://github.com/chili-lab/D-Intervention}.

\end{abstract}
\section{Introduction}
\label{sec:introduction}
As language models (LMs) continue to grow in complexity and capability, understanding and controlling their behavior has become increasingly critical~\cite{olah2015understanding,geva-etal-2021-transformer, yu2023whiteboxtransformerssparserate,geiger2024findingalignmentsinterpretablecausal}. Recent advances in interpretability research have highlighted the potential of model interventions–targeted modifications to model behavior during forward passes–as a promising approach to achieving this control~\cite{meng2023locatingeditingfactualassociations,conmy2023automatedcircuitdiscoverymechanistic, ghandeharioun2024patchscopesunifyingframeworkinspecting}. These interventions allow researchers to steer model behavior, often by manipulating representation in the model's latent space.

A core challenge in intervention-based methods lies in \textit{advancing from low-level control to high-level control}. Low-level control includes steering the model to output antonyms/synonyms, or binary labels (e.g., positive/negative sentiment). While these tasks served as foundational benchmarks for early intervention research, the field needs to address more complex, high-level behaviors~\cite{Zhang2023TowardsBP}. These tasks require interventions that operate at a deeper, more abstract level, capturing the intricate relationships and dependencies within the model's hidden representations. 

\begin{figure}[t]

\begin{center}
        \centering 
        \includegraphics[width=0.95\linewidth]{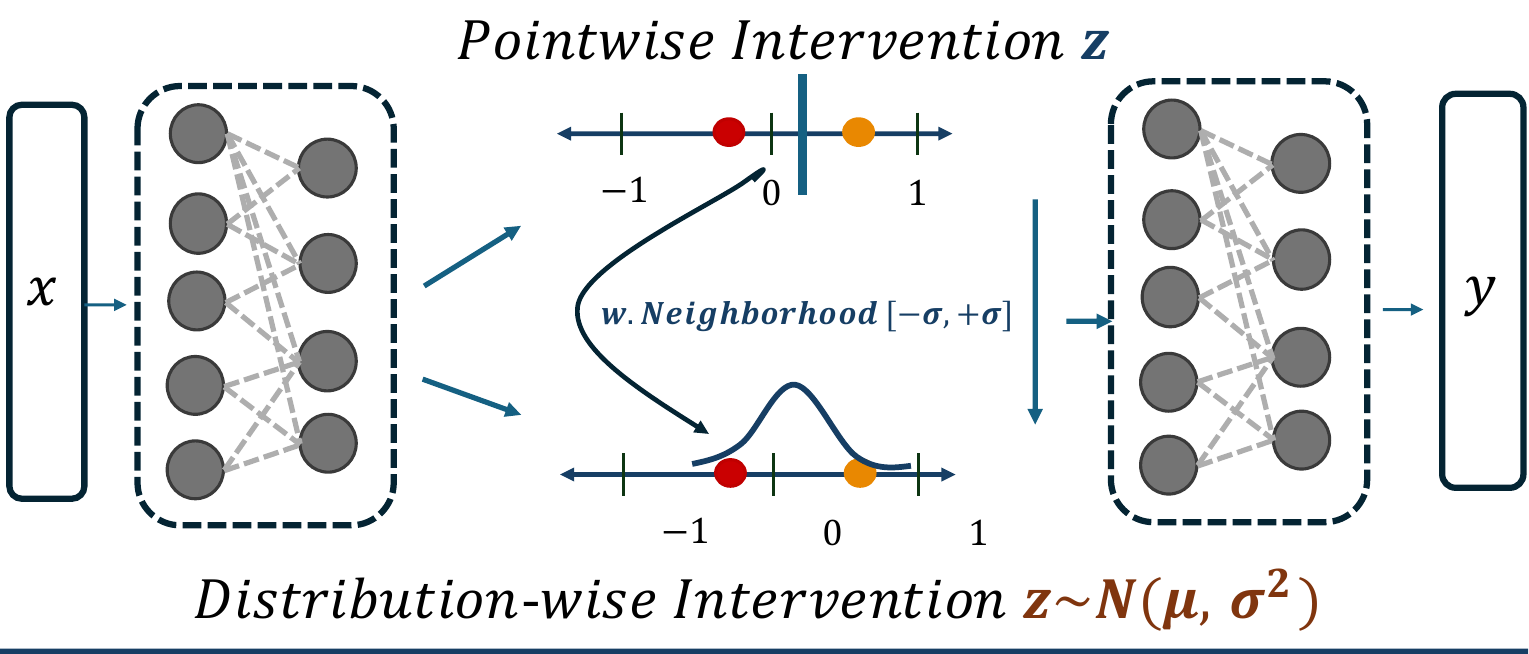}
        \caption{\textbf{Distribution-wise vs. Pointwise Intervention.} This is an intuitive yet effective adaptation, as previous research suggests that the concept space is continuous~\cite{gandikota2023conceptslidersloraadaptors}. }
        \label{fig:intro}
\end{center}
\vspace{-0.4cm}
\end{figure}

Learnable interventions, or \textit{representation fine-tuning}~\cite{wu2024reftrepresentationfinetuninglanguage,yin2024lofitlocalizedfinetuningllm}, have emerged as a promising solution. It enables more powerful high-level control, often outperforming parameter-efficient fine-tuning (PEFT) methods on tasks like commonsense question answering (QA), mathmetical reasoning and alignment tasks, while using 10x-100x fewer parameters~\cite{houlsby2019parameterefficienttransferlearningnlp, hu2021loralowrankadaptationlarge,hu-etal-2023-llm,liu2024doraweightdecomposedlowrankadaptation}. This highlights the potential to modify model behavior in a finer-grained manner.

A potential improvement for these methods is to explore an ideal ``concept space''. This space should be continuous, as previous research suggests that once an intervention vector is identified, its magnitude can be adjusted to control its effect~\cite{gandikota2023conceptslidersloraadaptors,zou2023representationengineeringtopdownapproach,turner2024steeringlanguagemodelsactivation}. This implies that even if an intervention is learned, its neighboring regions should also produce relevant effects. The key research question then arises—how to effectively explore this region.

As shown in Figure~\ref{fig:intro}, one intuitive method is to \textit{replace deterministic nodes with stochastic ones to directly learn the latent distribution}. A common way to achieve this is through reparameterization~\cite{kingma2022autoencodingvariationalbayes}, which enables efficient gradient-based optimization by decoupling randomness from the model's parameters. Specifically, the stochastic node is expressed as a differentiable function of a base distribution and the model's parameters. This allows for standard backpropagation while preserving the ability to sample from the learned distribution.

Building on these insights, we propose a simple yet effective improvement to enhance the exploration of the concept space in intervention-based methods. Specifically, we replace a deterministic node (neural network), with two separate networks. These networks independently learn the mean ($\mu$) and log variance ($\log\sigma^2$) of the latent distributions through gradient descent. This approach serves as a valid \textbf{drop-in replacement} for existing methods.

We follow the experimental setup of prior work~\cite{hu-etal-2023-llm,wu2024reftrepresentationfinetuninglanguage} and conduct comprehensive experiments, spanning eight commonsense reasoning benchmarks and seven mathematical reasoning benchmarks. We test performance on Llama-family models~\cite{touvron2023llama2openfoundation, dubey2024llama3herdmodels} under both layer-wise and all-layer configurations.

In our layer-wise experiments, we observed an intriguing performance gain: replacing deterministic nodes with stochastic counterparts in early layers significantly improved model performance, yielding gains of $+4\%$ to $+6\%$. Furthermore, we found that these gains strongly correlate with the learned variance of the stochastic nodes, suggesting that broader neighborhood exploration during training leads to enhanced performance.

We then apply insights from the intervention layer analysis to experiments across all layers. By varying the ratio of layers subjected to distribution-wise intervention, we identify an effective strategy: replacing the first few layers with stochastic nodes while retaining determinism in subsequent layers. This approach achieves consistent performance gains across all 15 benchmarks, along with significant improvements in robustness compared to pointwise intervention. These findings highlight the superiority of learned distribution-wise intervention over its pointwise counterpart.

\textbf{Our key contributions are:} 
\begin{itemize}[left=0pt, labelwidth=*, itemsep=0pt, topsep=1pt]
   \item We propose a simple yet effective intervention method by replacing deterministic nodes with stochastic ones, enabling better exploration of the concept space.
\item We demonstrate that this approach significantly improves model performance by intervening early layers.
\item We find that a mixed strategy—replacing only the first few layers with stochastic nodes while keeping the later layers deterministic—yields the best results in terms of both performance and robustness.
\end{itemize}

\section{Related Work}
\label{sec:related_work}
\paragraph{Reparametrization.}
The reparameterization trick is a widely used technique that enables neural networks to learn from sampling through gradient descent. It is commonly applied in variational autoencoders (VAE)~\cite{kingma2022autoencodingvariationalbayes,Tian2020LearningVM} and variational information bottlenecks (VIB)~\cite{tishby2000informationbottleneckmethod,tishby2015deeplearninginformationbottleneck,alemi2019deepvariationalinformationbottleneck} to help models learn latent distributions. VIB has been shown to effectively control the word embedding layer in language models~\cite{li2019specializingwordembeddingsfor,chen2020learningvariationalwordmasks} or downstream tasks~\cite{White2020ImprovingFO,Mahabadi2021VariationalIB,behjati-etal-2023-learning}. The term ``variational" in VAE/VIB refers to variational inference, where KL divergence is used to measure how close the approximation is to the true distribution. However, in our setting, we remove the KL loss term, making it no longer variational inference. Instead, it becomes a relaxed approach that allows language models to learn distributions without constraints.

\paragraph{Representation Fine-tuning and PEFT.}
Recently, representation fine-tuning has emerged as a method to provide high-level control over LMs' behavior. Two concurrent works, ReFT~\cite{wu2024reftrepresentationfinetuninglanguage} and LoFiT~\cite{yin2024lofitlocalizedfinetuningllm}, address this issue from different perspectives. ReFT builds on the theory of distributed alignment search (DAS)~\cite{geiger2021causalabstractionsneuralnetworks,geiger2024findingalignmentsinterpretablecausal}, demonstrating that orthogonal properties in latent subspaces are crucial for enabling task-independent control \textit{between transformer blocks}. LoFiT, on the other hand, draws inspiration from traditional intervention research, focusing on a paradigm of localization and editing~\cite{meng2023locatingeditingfactualassociations,stolfo-etal-2023-mechanistic}. Both approaches share a common goal: bringing interpretability research into the realm of high-level behavior control, comparable to parameter-efficient fine-tuning (PEFT) methods like adapters~\cite{houlsby2019parameterefficienttransferlearningnlp,pfeiffer-etal-2020-mad,pmlr-v139-fu21a,hu-etal-2023-llm,zhang2024llamaadapterefficientfinetuninglanguage} and LoRA~\cite{hu2021loralowrankadaptationlarge,liu2024doraweightdecomposedlowrankadaptation,zhang2024autoloraautomaticallytuningmatrix}. 
Their results show that learnable interventions can match or even outperform PEFT methods, often with significantly fewer resources. Our method aligns more closely with ReFT, as it involves learning latent distributions between encoder and decoder (i.e., between transformer blocks). Therefore, we will primarily evaluate our stochastic vs. deterministic node approach within the ReFT framework.

\paragraph{Intervention-based interpretability.}
Intervention-based interpretability focuses on manipulating a model’s internal states to understand how LMs represent various behaviors~\cite{subramani-etal-2022-extracting,zou2023representationengineeringtopdownapproach,turner2024steeringlanguagemodelsactivation,li2024inferencetimeinterventionelicitingtruthful}. By intervening on specific linear subspaces of latent representations, researchers have revealed that human-interpretable concepts~\cite{rumelhart1986parallel}, such as linguistic features (e.g., gender, number)~\cite{hewitt-manning-2019-structural,lasri-etal-2022-probing,wang2022interpretabilitywildcircuitindirect, hanna-etal-2023-language,arora2024causalgymbenchmarkingcausalinterpretability,huang2024ravelevaluatinginterpretabilitymethods} and logical reasoning, are often encoded linearly within these models~\cite{wu2024interpretabilityscaleidentifyingcausal,deng2024languagemodelssymboliclearners,gurarieh2025enhancingautomatedinterpretabilityoutputcentric}. Techniques like concept erasure and subspace interventions have been instrumental in disentangling such attributes~\cite{belrose2023leaceperfectlinearconcept,ravfogel2024linearadversarialconcepterasure}, enabling targeted modifications to improve model fairness, explainability, and task performance~\cite{nanda2023emergentlinearrepresentationsworld,park2024linearrepresentationhypothesisgeometry}. These works showcase that the representation space of language models encodes rich, structured information that is highly relevant to tasks, enabling more effective and targeted interventions.
\section{Preliminary}
\label{sec:preliminary}
We now first introduce the background of our intervention methods.\footnote{In this work, we primary focus on learnable intervention between layers (i.e. transformer block).} First, we outline the formulation of transformer-based decoder LMs and their layer-wise hidden representations. Then, we will provide a unified view of intervention from an information-theoretic perspective.

\subsection{Transformer Architecture}
Transformer-based autoregressive LMs~\cite{vaswani2023attentionneed} aim to predict the probability of a sequence of tokens. Let \(X = \{x_1, x_2, \ldots, x_n\}\) denote the input sequence, where each \(x_i\) represents a token in the sequence. Let $Y = \{y_1, y_2, ..., y_m\}$ denote the output sequence. Overall, the goal of next-token prediction in language modeling can be formally represented as estimating $P(Y|X)$. 

The hidden representations at each layer of the model act as latent variables \(Z\), encoding intermediate abstractions that bridge \(X\) and \(Y\):
\[
Z^{(l)} = \{z^{(l)}_1, z^{(l)}_2, \ldots, z^{(l)}_n\}, \quad l = 1, 2, \ldots, L,
\]
where \(l\) indexes the layers of the language model, and \(z^{(l)}_i\) represents the hidden state of token \(x_i\) at layer \(l\).

\subsection{Layer-Wise Representation Transformation}
Each layer in a language model is designed to transform the latent representations \(Z^{(l)}\) using contextual information from neighboring tokens. These transformations can be expressed as:
\[
Z^{(l+1)} = Attn(Z^{(l)}) + FFN(Attn(Z^{(l)})),
\]
where \(FFN\) represents a feed-forward networks (FFN) with the input from \(Z^{(l)}\), and \(Attn(\cdot)\) represents the self-attention module within the transformer block.

\subsection{Intervention in Language Models}
In this work, we provide an \textit{information-theoretical view} of model intervention. In established information theory research, a common method for estimating mutual information is achieved by inserting an auxiliary network (e.g., a variational autoencoder) at a specific layer.
 
Interestingly, in intervention research, interventions in a language model are formalized as a function $f_{\phi}$, parameterized by $\phi$,  that transforms hidden representations $Z^{(l)}$ at layer $l$ to modified representations $\hat{Z}^{(l)}$:
\begin{equation}
\vspace{-0.2cm}
    \hat{Z}^{(l)} = f_{\phi}(Z^{(l)}).
\end{equation}

We observe a \textbf{connection} here: the auxiliary network used in information theory for mutual information estimation can be viewed as a specific form of intervention. It represents a special case where the transformation function $f_{\phi}$ is a learnable variational autoencoder.

In practice, the goal of intervention can be viewed as improving downstream task performance. For learnable interventions, this translates to minimizing the cross-entropy (CE) loss between the predicted and true outputs:
\begin{equation}
\mathcal{L}_{CE} = -\mathbb{E}_{(X,Y)} \left[ \log P(Y | f_{\phi}(Z^{(l)})) \right].
\end{equation}
The cross-entropy loss can be directly recognized as the conditional entropy:
\begin{equation}
\mathcal{L}_{CE} = -\mathbb{E}_{(X,Y)} \left[ \log P(Y | f_{\phi}(Z^{(l)})) \right] = H(Y | f_{\phi}(Z^{(l)})).
\end{equation}
The mutual information between $Y$ and the transformed representations is defined as:
\begin{equation}
I(Y; f_{\phi}(Z^{(l)})) = H(Y) - H(Y | f_{\phi}(Z^{(l)})).
\end{equation}
Rearranging this expression:
\begin{equation}
H(Y | f_{\phi}(Z^{(l)})) = H(Y) - I(Y; f_{\phi}(Z^{(l)})).
\end{equation}
Substituting this into the cross-entropy loss:
\begin{equation}
\mathcal{L}_{CE} = H(Y | f_{\phi}(Z^{(l)})) = H(Y) - I(Y; f_{\phi}(Z^{(l)})).
\end{equation}
Given that $H(Y)$ is constant with respect to the intervention parameters $\phi$, minimizing the cross-entropy loss is equivalent to maximizing the mutual information:
\begin{equation}
\arg\min_{\phi} \mathcal{L}_{CE} \equiv \argmax_{\phi} I(Y; f_{\phi}(Z^{(l)})).
\end{equation}
This formulation captures the fundamental goal of interventions: \textit{to transform the internal representations such that they become maximally informative about the target output}. Conceptually, this optimization seeks the intervention that best preserves and amplifies the signal relevant to the task while potentially filtering out irrelevant information.

\section{Distribution-Wise Intervention}
\setcounter{footnote}{0}
\label{sec:distribution-level-intervention}
In this section, we first introduce the motivation for distribution-wise control and provide a detailed description of the improvements, specifically replacing a deterministic node with a stochastic node that can learn from  sampling.

\subsection{Motivation}
Many previous studies have found that the effect of an intervention can be controlled by adjusting its magnitude~\cite{gandikota2023conceptslidersloraadaptors,turner2024steeringlanguagemodelsactivation,han2024wordembeddingssteerslanguage}. The intervention effect should not be limited to a single point; rather, its surrounding neighborhood must also exhibit relevant effects. This suggests that the impact of an intervention propagates across related regions. A useful analogy is the transition from autoencoders (AE) to variational autoencoders (VAE). VAEs replace deterministic nodes with stochastic sampling, allowing the model to learn latent distributions directly~\cite{kingma2022autoencodingvariationalbayes}. 
We explore applying the technique to intervention research, investigating whether they can help learn better interventions.

\subsection{Stochastic Intervention Reparameterization}
To effectively learn distributions through stochastic nodes, we employ the reparameterization trick~\cite{kingma2022autoencodingvariationalbayes}. This technique enables gradient-based optimization through sampling by reformulating the random sampling process as a deterministic function of the distribution parameters and an auxiliary noise variable.

Consider a simple deterministic MLP layer that transforms input representation $Z$ through:
\begin{equation}
   {\hat{Z}} = \text{MLP}(Z) = W^TZ + b.
\end{equation}
We replace this with a stochastic layer that learns a distribution $\mathcal{N}(\boldsymbol{\mu}, \boldsymbol{\sigma}^2)$. Instead of directly sampling from this distribution, which would break gradient flow, we reparameterize the sampling process:
\begin{align}
    \boldsymbol{\mu} &= \text{MLP}_{\mu}(Z), \\
    \log\sigma^{2}
    &= \mathrm{MLP}_{\log\sigma^{2}}(Z), \\
    \boldsymbol{\sigma}    &= \exp\bigl(\tfrac12\,\log\sigma^{2}\bigr), \\
    \boldsymbol{\epsilon} &\sim \mathcal{N}(0, I), \\
    \hat{Z} &= \boldsymbol{\mu} + \boldsymbol{\sigma} \odot \boldsymbol{\epsilon}.
\end{align}
Here, $\text{MLP}_{\mu}$ and $\text{MLP}_{\log\sigma^{2}}$ learn the distribution parameters, while $\odot$ denotes element-wise multiplication. The stochasticity comes from $\boldsymbol{\epsilon}$ (random noise), and it allows gradients to flow through $\boldsymbol{\mu}$ and $\boldsymbol{\sigma}$ during backpropagation while maintaining the stochastic nature of the transformation through $\boldsymbol{\epsilon}$. 
 
\subsection{Training Objective}
 Given a frozen base language model $\mathcal{M}$ and trainable stochastic intervention layers $\{\mathcal{I}_l\}_{l=1}^L$ inserted between transformer blocks, we minimize the cross-entropy loss over the next-token prediction task:
\[
\mathcal{L} = -\mathbb{E}_{(X,Y)} \left[ \log P_{\mathcal{M} \circ \mathcal{I}}(Y|f_{\phi}(Z^{(l)})) \right],
\]
where $\mathcal{M} \circ \mathcal{I}$ denotes the composed system of the frozen language model with our stochastic intervention layers. During training, gradients flow through the reparameterized stochastic networks  back to the learnable parameters $\{\phi_{\mu}^{(l)}, \phi_{\sigma}^{(l)}\}_{l=1}^L$ of the $\text{MLP}_{\mu}$ and $\text{MLP}_{\sigma}$ networks..

\subsection{Model-Specific Clamping}
To address numerical instability issues arising from large sampling variances, we introduce model-specific clamping based on the weight distributions of the target language model. Given a language model $\mathcal{M}$ with intervention at layer $l$, we define the clamping boundaries using the statistics of adjacent layer weights.
Let $W^{(l)}$ and $W^{(l+1)}$ denote the weight matrices before and after the intervention layer respectively. We define the clamping bounds as:
\begin{align}
v_{\text{min}} &= \min(\min(W^{(l)}), \min(W^{(l+1)})) ,\\
v_{\text{max}} &= \max(\max(W^{(l)}), \max(W^{(l+1)})).
\end{align}
This model-specific clamping ensures that the interventions remain within the natural range of the model's weight distributions, helping to maintain stability while preserving the model's learned representations. The bounds are computed once before training and remain fixed throughout the intervention process.

\begin{tcolorbox}[colback=gray!10, colframe=gray!50, title={Capturing Uncertainty with Stochastic Layers}]
During training, optimizing both the mean and variance of the learned distribution enables the model to effectively capture uncertainty in the intervention space. This stochastic approach offers two key advantages: 
\begin{itemize}[left=0pt, labelwidth=*, itemsep=0pt, topsep=1pt]
   \item It facilitates exploration of the intervention neighborhood through sampling.  
\item It allows the model to learn and represent uncertainty in intervention effects, improving its robustness and adaptability.  
\end{itemize}
\end{tcolorbox}
\section{Experiment Setup}
\label{sec:experiment-setup}
To evaluate our distribution-level intervention methods compared with pointwise intervention, we evaulate our methods on more than ten datasets with full combination of different hyperpameter tuning. Generally, we follow the standard setup of previous SOTA methods like ReFT~\cite{wu2024reftrepresentationfinetuninglanguage}, and our codebase is built on pyenve~\cite{wu-etal-2024-pyvene}. ReFT's evaluation framework is also derived from prior work like~\cite{hu-etal-2023-llm,liu2024doraweightdecomposedlowrankadaptation,wu2024advancingparameterefficiencyfinetuning}. Similar to these previous work, we evaluate Llama-series model~\cite{touvron2023llamaopenefficientfoundation,touvron2023llama2openfoundation,dubey2024llama3herdmodels} ranging from Llama-7B/13B to Llama-3-8B. We conducted all experiments using a single NVIDIA RTX A6000 GPU with mixed precision (\texttt{bfloat16}) enabled. 

Our evaluation is divided into two parts: (i) \textit{layer-wise setting} and (ii) \textit{all-layer setting}. First, we analyze layer-wise control by experimenting with different types of interventions. We then explore how replacing these interventions with distribution-level controls affects performance. Finally, we evaluate the interventions in an all-layer setting and compare with the results in previous literature.

\subsection{Baselines}

For the \textit{layer-wise setting}, in addition to RED~\cite{wu2024advancingparameterefficiencyfinetuning} and ReFT, we also include simple MLP and SwiGLU~\cite{shazeer2020gluvariantsimprovetransformer} as baselines to evaluate the impact of distribution-level intervention.\footnote{We denote the distribution-level variants of \textit{ReFT} as \textit{D-ReFT}, and similar notation applies to other methods} Concrete formats are provided below.

For the \textit{all-layer setting}, we perform a comparative analysis of ReFT and previous parameter-efficient fine-tuning (PEFT) methods. These include: Prefix-tuning~\cite{li2021prefixtuningoptimizingcontinuousprompts},  RED~\cite{wu2024advancingparameterefficiencyfinetuning},  LoRA~\cite{hu2021loralowrankadaptationlarge}, DoRA~\cite{liu2024doraweightdecomposedlowrankadaptation} and  ReFT~\cite{wu2024reftrepresentationfinetuninglanguage}.

\begin{tcolorbox}[colback=gray!10, colframe=gray!50, title={Intervention Functions}]
\label{tbox:intervention-function}
Pointwise intervention function $f_\phi$:
\begin{itemize}[left=0pt, labelwidth=*, itemsep=0pt, topsep=1pt]

    \item \textbf{MLP}: \( \hat{Z} = W^T Z + b \)
    
    \item \textbf{RED}: \( \hat{Z} = W \odot Z + b \)
    
    \item \textbf{SwiGLU}: \( \hat{Z} = (W \odot Z + b) \odot GELU(Z) \)
    
    \item \textbf{ReFT}: \( \hat{Z} = Z + R(W^T Z + b - R^T Z) \)

\end{itemize}
Distribution-wise intervention function $f'_\phi$:
\begin{itemize}[left=0pt, labelwidth=*, itemsep=0pt, topsep=1pt]
\item \textbf{D-MLP}: \( \hat{Z} = \boldsymbol{\mu} + \boldsymbol{\sigma} \odot \boldsymbol{\epsilon} \)

\item \textbf{D-RED}: \( \hat{Z} = \boldsymbol{\mu} + \boldsymbol{\sigma} \odot \boldsymbol{\epsilon} \)

\item \textbf{D-SwiGLU}: \( \hat{Z} = \boldsymbol{\mu} \odot GELU(Z) + \boldsymbol{\sigma} \odot \boldsymbol{\epsilon} \)

\item \textbf{D-ReFT}: \( \hat{Z} = Z + R(\boldsymbol{\mu} + \boldsymbol{\sigma} \odot \boldsymbol{\epsilon} - R^T Z) \)
\end{itemize}
\end{tcolorbox}

\subsection{Benchmark}

We evaluate our methods on seven commonsense reasoning benchmarks and seven arithmetic reasoning benchmark. 

For commonsense reasoning, we have BoolQ~\cite{clark2019boolqexploringsurprisingdifficulty}, PIQA~\cite{bisk2019piqareasoningphysicalcommonsense}, SIQA~\cite{sap2019socialiqacommonsensereasoningsocial}, HellaSwag ~\cite{zellers2019hellaswagmachinereallyfinish},
WinoGrande~\cite{sakaguchi2019winograndeadversarialwinogradschema}, ARC-e, ARC-c~\cite{clark2018thinksolvedquestionanswering} and OBQA~\cite{mihaylov2018suitarmorconductelectricity}. The input format is multi-choice QA, given a context or a question with multiple answer choices. The output is as simple as the selected choice, without CoT rationales. 

For arithmetic reasoning, we have AddSub~\cite{hosseini-etal-2014-learning}, SingleEQ~\cite{koncel-kedziorski-etal-2015-parsing}, MultiArith~\cite{roy2016solvinggeneralarithmeticword},  AQuA~\cite{ling2017programinductionrationalegeneration}, GSM8K~\cite{cobbe2021trainingverifierssolvemath}, MAWPS~\cite{koncel-kedziorski-etal-2016-mawps}, and SVAMP~\cite{patel-etal-2021-nlp}.
For the arithmetic reasoning benchmarks, CoT rationale are given before the final answer. 

For all benchmarks, We use the same prompt template as in~\citet{hu-etal-2023-llm,wu2024reftrepresentationfinetuninglanguage}. We also remove leading and trailing whitespace in the dataset.

\subsection{Hyperparameter Tuning}
For the commonsense reasoning benchmark, we train the model using the \texttt{Commonsense170K} dataset. For arithmetic reasoning benchmarks, we use the \texttt{Math10K} dataset. These datasets are combined training sets from their original benchmarks. We use a portion of the training set from GSM8K as a development set to tune the best hyperparameters and apply this set of hyperparameters to report the test scores. We do not optimize directly on the test set. This setting is the same as that used by~\citet{wu2024reftrepresentationfinetuninglanguage}.

Key parameters include the \textit{intervention layer} ($l$), \textit{noise scale} ($\epsilon$), \textit{subspace rank} ($r$), \textit{intervention position} ($p$), \textit{batch size} ($bs$), \textit{training epochs} ($e$), and \textit{learning rate} ($lr$). These parameters are tuned on the development set, but an ablation study is not included in the main text. Detailed values are provided in Appendix~\ref{appendix:hyperparameter}.

\section{Layer-Wise Intervention}
\label{sec:laywe-wise-control}
Previous work often reports results using the all-layer setting (i.e., interventions applied across all layers). In this study, we first conduct layer-wise ablation experiments to identify where the performance gains come from. 
\begin{figure*}[t]
    \centering
    \subfigure[MLP]{
        \includegraphics[width=0.23\textwidth]{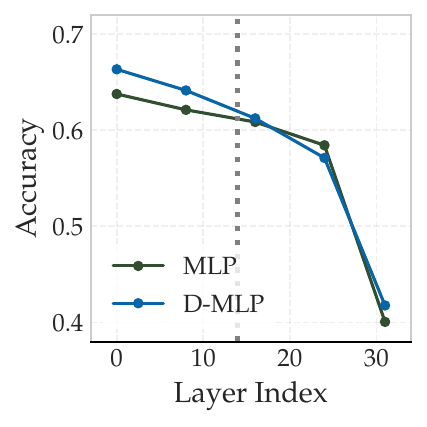} %
    }
    \subfigure[SwiGLU]{
        \includegraphics[width=0.23\textwidth]{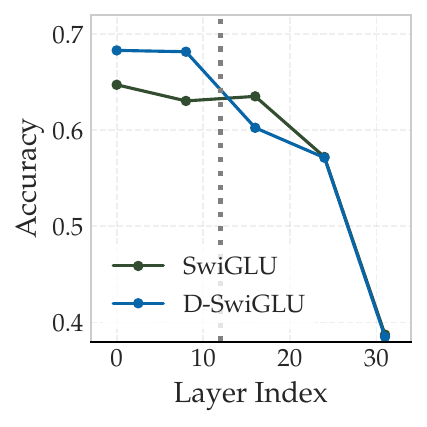} %
    }
    \subfigure[RED]{
        \includegraphics[width=0.23\textwidth]{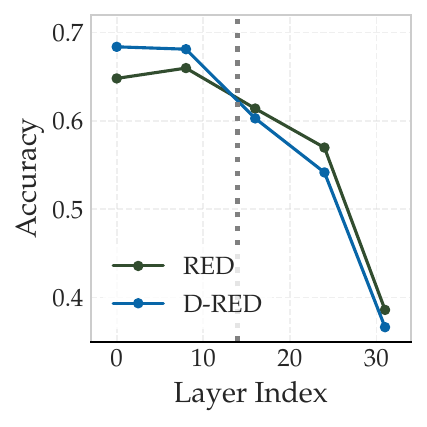} %
    }
    \subfigure[ReFT]{
        \includegraphics[width=0.23\textwidth]{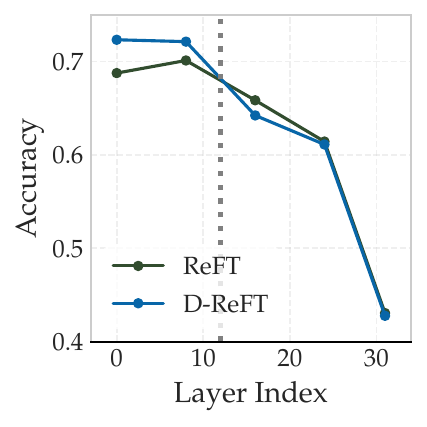} %
    }
    \caption{\textbf{Performance of different layer-wise D-interventions with respect to the intervention layer}. We report the average score of Llama-3-8B on seven arithmetic benchmarks: AddSub, SingleEQ, MultiArith, AQuA, GSM8K, MAWPS, and SVAMP. Notably, D-intervention at early layers yields the best performance, highlighting a significant discrepancy across layers.}
    \label{fig:intervention-layer}
\end{figure*}
\subsection{Abalation Study: D-Intervention Layer $l$}
We evaluate distribution-level controllability in arithmetic reasoning across seven benchmarks: AddSub, SingleEQ, MultiArith, AQuA, GSM8K, MAWPS, and SVAMP. All methods are tuned on the development set, and we report the average results over three runs with different seeds.

\paragraph{Results.}
As shown in Figure~\ref{fig:intervention-layer}, ReFT maintains superior performance compared to RED, SwiGLU, and MLP across all four methods. Both standard intervention and its distribution-level variant reveal a consistent pattern: deeper layers lead to significant performance declines. The accuracy drops from 0.7 to approximately 0.4 as interventions move to later layers, demonstrating that later-layer interventions are more challenging than those in earlier layers. 

This follows from data processing inequality (see appendix~\ref{appendix:dpi}), which states that deep layers cannot recover information lost in earlier layers. Thus, intervention should be applied early to preserve useful information before transformations degrade it. Across MLP, SwiGLU, RED, and ReFT, their \textit{D-variants consistently boost accuracy by around +4\% in early layers}, which is higher than the improvements from LoRA $\rightarrow$ DoRA $\rightarrow$ ReFT~\cite{liu2024doraweightdecomposedlowrankadaptation,wu2024reftrepresentationfinetuninglanguage}. This highlights that distribution-level intervention is an effective improvement when applied early in the network.

\subsection{Ablation Study: D-Intervention Noise $\boldsymbol{\epsilon}$}
We then conduct an ablation study on the stochastic node $\boldsymbol{\epsilon}$. In our method, $\boldsymbol{\epsilon} \odot \boldsymbol{\sigma}$ plays a key role in transforming a deterministic node into a stochastic one to learn the distribution. Here, $\boldsymbol{\sigma}$ represents the standard deviation matrix of the latent variable $Z$, while $\boldsymbol{\epsilon}$ controls the magnitude of this learning effect. Adjusting $\boldsymbol{\epsilon}$ allows us to explore the true distribution or the ideal concept space.

\begin{figure}[t]

\begin{center}
        \centering 
        \includegraphics[width=0.95\linewidth]{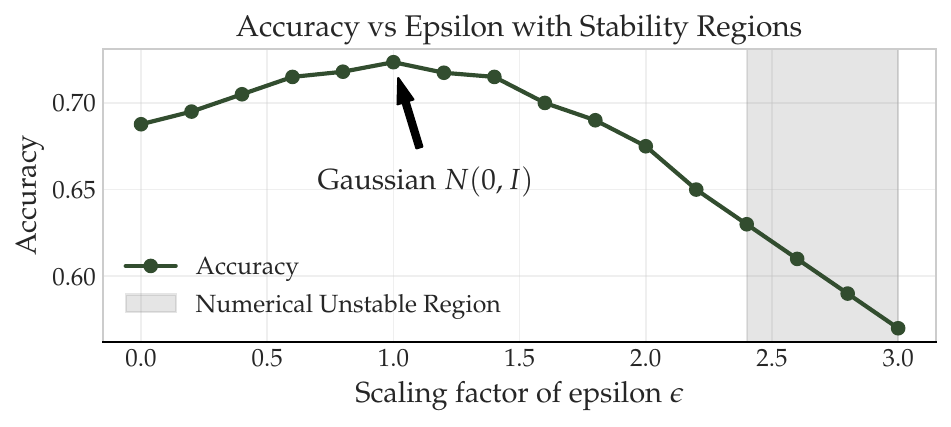}
        \caption{\textbf{Accuracy with different scaling factors for \( \boldsymbol{\epsilon} \) in D-ReFT.} When the scaling factor for \( \boldsymbol{\epsilon} = 0\), the method reduces to the original ReFT. For scaling factors of \( \boldsymbol{\epsilon} > 2.4 \), D-ReFT enters a region prone to numerical instability due to large  variance.}
        \label{fig:epsilon-acc}
\end{center}
\vspace{-0.4cm}
\end{figure}

\paragraph{Results.}
\vspace{-0.3cm}
We vary the scaling factor $\lambda$ applied to $\boldsymbol{\epsilon}$ from $0$ to $3.0$, using a step size of 0.2, in the D-ReFT setting. As shown in Figure~\ref{fig:epsilon-acc}, performance improves from ReFT to D-ReFT with (scaling factor of $\boldsymbol{\epsilon=1}$) where the best results are achieved. This showcase that default setting with $\boldsymbol{\epsilon} \sim \mathcal{N}(0, I)$ still has stable performance. However, increasing $\boldsymbol{\epsilon}$ further introduces higher variance, making training more difficult and causing numerical instability.

\begin{figure}[t]
\vspace{-0.4cm}
    \centering
    \subfigure[$Acc=0.688$]{
        \includegraphics[width=0.15\textwidth]{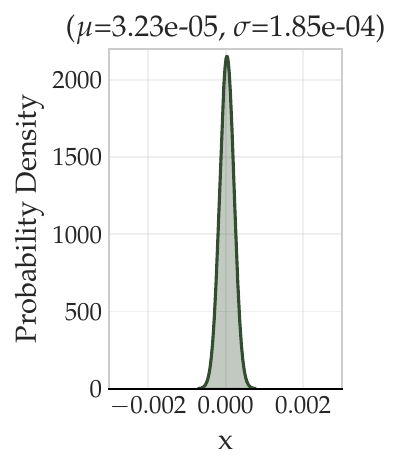} %
    }
    \hspace{-0.2cm}
    \subfigure[$Acc=0.723$]{
        \includegraphics[width=0.15\textwidth]{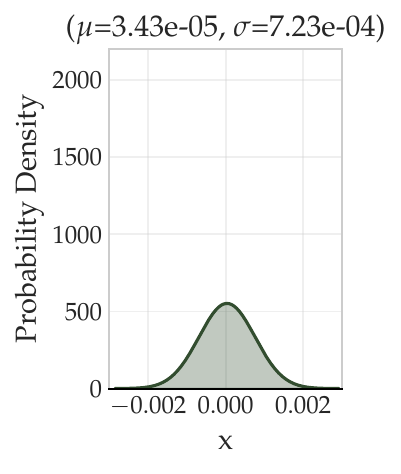} %
    }
    \hspace{-0.2cm}
     \subfigure[$Acc=0.694$]{
        \includegraphics[width=0.15\textwidth]{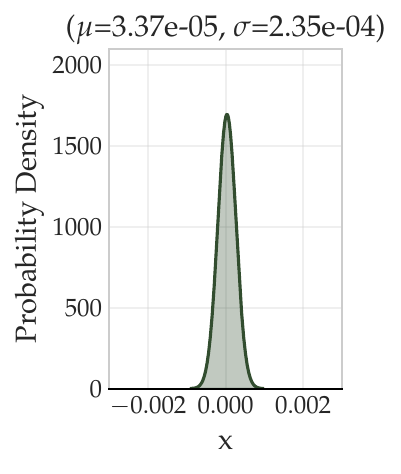} %
    }
    \caption{\textbf{Correlation between accuracy and std $\boldsymbol{\sigma}$}. The scaling factors $\lambda$ of $\boldsymbol{\epsilon}$ are 0.2, 1.0, and 1.8 for subfigures (a), (b), and (c).}
    \label{fig:gaussian-pdf}
\vspace{-0.4cm}
\end{figure}
We explore the correlation between accuracy and the standard deviation of the learned distribution. Using all test set samples, we compute the average accuracy for different standard deviations. As shown in Figure~\ref{fig:gaussian-pdf}, accuracy and standard deviation exhibit a positive correlation: \textit{distributions with higher standard deviation tend to perform better, while lower standard deviation is associated with worse performance.} This suggests that distribution-level interventions are effective because they explore the neighborhood region of pointwise interventions.

\begin{table*}[t]
    \centering
    \caption{Performance comparison of LLaMA-1 7B/13B, Llama-2 7B and Llama-3 8B against existing PEFT methods on eight commonsense reasoning datasets.
    $^*$Performance results of all baseline methods are taken from \citet{wu2024reftrepresentationfinetuninglanguage,liu2024doraweightdecomposedlowrankadaptation}. \textbf{D-ReFT}\textsubscript{25\%}, for example in Llama-3-8B, represents replace first 8 layers to D-ReFT and remain the left 24 layers as ReFT intervention.}
    \label{tab:commonsense_result}
    \adjustbox{max width=\textwidth}{
    \begin{tabular}{llrrrrrrrrrl}
        \toprule
        \multirow{2}{*}{\textbf{Model}} & \multirow{2}{*}{\textbf{PEFT}} & \multirow{2}{*}{\textbf{Params} (\%)} & \multicolumn{8}{c}{\textbf{Accuracy} ($\uparrow$)} \\
        \cmidrule{4-12}
         \rowcolor{gray!15}
        & & & \textbf{BoolQ} & \textbf{PIQA} & \textbf{SIQA} & \textbf{HellaS.} & \textbf{WinoG.} & \textbf{ARC-e} & \textbf{ARC-c} & \textbf{OBQA} & \textbf{Avg.} \\
        \midrule
        ChatGPT$^*$ & --- & --- & 73.1 & 85.4 & 68.5 & 78.5 & 66.1 & 89.8 & 79.9 & 74.8 & 77.0 \\ \cmidrule{1-12}
        \multirow{8}{*}{LLaMA-7B} & PrefT$^*$ & 0.039\% & 64.3 & 76.8 & 73.9 & 42.1 & 72.1 & 72.9 & 54.0 & 60.6 & 64.6 \\
        & Adapter\textsuperscript{S}$^*$ & 1.953\% & 63.0 & 79.2 & 76.3 & 67.9 & 75.7 & 74.5 & 57.1 & 72.4 & 70.8 \\
        & Adapter\textsuperscript{P}$^*$ & 3.542\% & 67.9 & 76.4 & 78.8 & 69.8 & 78.9 & 73.7 & 57.3 & 75.2 & 72.3 \\
        & LoRA$^*$ & 0.826\% & 68.9 & 80.7 & 77.4 & 78.1 & 78.8 & 77.8 & 61.3 & 74.8 & 74.7 \\
        & DoRA$^*$ & 0.838\% & 68.5 & 82.9 & 79.6 & 84.8 & 80.8 & 81.4 & 65.8 & 81.0 & 78.1 \\ 
         & ReFT$^*$ & 0.031\% & 69.3 & 84.4 & 80.3 & 93.1 & 84.2 & 83.2 & 68.2 & 78.9 & 80.2 \\ \cmidrule{2-12}
          \rowcolor{gray!15}
         & \textbf{D-ReFT}\textsubscript{25\%} (Ours) & 0.046\% & \textbf{72.1} & \textbf{87.4} & \textbf{81.1} & \textbf{93.7} & \textbf{85.4} & \textbf{84.7} & \textbf{71.7} & \textbf{80.4} & \textbf{82.2} \\
        \midrule
        \multirow{8}{*}{LLaMA-13B} & PrefT$^*$ & 0.031\% & 65.3 & 75.4 & 72.1 & 55.2 & 68.6 & 79.5 & 62.9 & 68.0 & 68.4 \\
        & Adapter\textsuperscript{S}$^*$ & 1.586\% & 71.8 & 83.0 & 79.2 & 88.1 & 82.4 & 82.5 & 67.3 & 81.8 & 79.5 \\
        & Adapter\textsuperscript{P}$^*$ & 2.894\% & 72.5 & 84.9 & 79.8 & 92.1 & 84.7 & 84.2 & 71.2 & 82.4 & 81.5 \\
        & LoRA$^*$ & 0.670\% & 72.1 & 83.5 & 80.5 & 90.5 & 83.7 & 82.8 & 68.3 & 82.4 & 80.5 \\
        & DoRA$^*$ & 0.681\% & 72.4 & 84.9 & 81.5 & 92.4 & 84.2 & 84.2 & 69.6 & 82.8 & 81.5 \\ 
         & ReFT$^*$ & 0.025\% & 72.1 & 86.3 & 81.8 & 95.1 & 87.2 & 86.2 & \textbf{73.7} & 84.2 & 83.3 \\ \cmidrule{2-12}
          \rowcolor{gray!15}
          & \textbf{D-ReFT}\textsubscript{25\%} (Ours) & 0.037\% & \textbf{74.3} & \textbf{87.1} & \textbf{83.3} & \textbf{95.2} & \textbf{89.3} & \textbf{87.1} & 73.6 & \textbf{85.9}& \textbf{85.1} \\
        \midrule
        \multirow{5}{*}{Llama-2 7B} 
        & LoRA$^*$ & 0.826\% &  69.8 & 79.9 & 79.5 & 83.6 & 82.6 & 79.8 & 64.7 & 81.0 & 77.6 \\
        & DoRA$^*$ & 0.838\% &  \textbf{71.8} & 83.7 & 76.0 & 89.1 & 82.6 & 83.7 & 68.2 & 82.4 & 79.7 \\ 
            & ReFT$^*$ & 0.031\% & 71.1 & 83.8 & 80.8 & \textbf{94.3} & 84.5 & 85.6 & 72.2 & 82.3 & 81.8 \\ \cmidrule{2-12}
             \rowcolor{gray!15}
             & \textbf{D-ReFT}\textsubscript{25\%} (Ours) & 0.046\% & 71.3 & \textbf{86.7} & \textbf{81.8} & 94.1 & \textbf{87.3} & \textbf{86.1} & \textbf{73.0} & \textbf{84.2} & \textbf{83.6} \\
        \midrule
        \multirow{5}{*}{Llama-3 8B} 
        & LoRA$^*$ & 0.700\% & 70.8 & 85.2 & 79.9 & 91.7 & 84.3 & 84.2 & 71.2 & 79.0 & 80.8 \\
        & DoRA$^*$ & 0.710\% & 74.6 & 89.3 & 79.9 & 95.5 & 85.6 & 90.5 & 80.4 & 85.8 & 85.2 \\ 
            & ReFT$^*$ & 0.026\% & 75.1 & 90.2 & 82.0 & \textbf{96.3} & 87.4 & 92.4 & 81.6 & 87.5 & 86.6 \\ \cmidrule{2-12}
             \rowcolor{gray!15}
        & \textbf{D-ReFT}\textsubscript{25\%} (Ours) & 0.039\% & \textbf{78.3} & \textbf{93.4} & \textbf{83.7} & 96.1 & \textbf{89.7} & \textbf{94.9} & \textbf{83.1} & \textbf{89.4} & \textbf{89.1} \\
        \bottomrule\\
    \end{tabular}
    }
\vspace{-0.3cm}
\end{table*}

\section{All-layer Intervention}
\label{sec:all-layer-control}

\vspace{-0.1cm}
We evaluate D-ReFT’s performance across all network layers, focusing on how layer allocation impacts results. Since the strongest improvements occur in early layers, we conduct an ablation study to test a mixed strategy of pointwise and distribution-level interventions. Specifically, we vary how many early layers are replaced with D-ReFT. We compare four configurations: D-ReFT\textsubscript{25\%}, D-ReFT\textsubscript{50\%}, D-ReFT\textsubscript{75\%}, and D-ReFT\textsubscript{100\%}, where the subscript indicates the percentage of early layers replaced with D-ReFT.
\subsection{Commonsense Reasoning}
Table~\ref{tab:commonsense_result} shows the performance comparison on eight different commonsense reasoning benchmarks. The results show that D-ReFT\textsubscript{25\%} consistently outperforms its original pointwise version ReFT and existing PEFT methods across all LLaMA models. These results highlight the effectiveness of D-ReFT in enhancing reasoning performance while maintaining efficiency.
Table~\ref{tab:mixed-strategy-REFT} presents the results of mixed strategies for ReFT and D-ReFT. We observe a consistent trend: applying D-ReFT to the first 25\% of layers and ReFT to the remaining layers yields the best performance, while applying D-ReFT to greater than some threshold (i.e., introducing too much randomness) during training would make the model hard to converge. This suggests that a mixed strategy—using stochastic nodes in early layers and deterministic nodes in later layers—is optimal for language models.
\begin{table}[t]
\vspace{-0.4cm}
    \centering
    \caption{Accuracy (\%) in the commonsense tasks when replacing ReFT to D-ReFT intervention for the top $\%$ layers.}
    \label{tab:mixed-strategy-REFT}
    \vspace{+0.2cm}
    \resizebox{0.5\textwidth}{!}{
    \begin{tabular}{lccccc}
        \toprule
        Model $\mathcal{M}$ & ReFT & \textbf{D-ReFT}\textsubscript{25\%} & D-ReFT\textsubscript{50\%} & D-ReFT\textsubscript{75\%} & D-ReFT\textsubscript{100\%} \\
        \midrule
        LLaMA-7B & 80.2 & 82.2 & 81.0 & 79.4 & 79.3 \\
        LLaMA-13B & 83.3 & 85.1 & 83.9 & 83.0 & 82.6 \\
        LLaMA-2-7B & 81.8 & 83.6 & 82.0 & 81.9 & 80.9 \\
        LLaMA-3-8B & 86.6 & 89.1 & 86.7 & 85.9 & 85.1 \\
        \bottomrule
    \end{tabular}}
    \vspace{-0.4cm}
\end{table}

\begin{table*}[t]
    \centering
        \caption{Performance comparison of LLaMA-1 7B, Llama-2-7B and Llama-8B against existing PEFT methods on four arithmetic reasoning datasets. \textbf{D-ReFT}\textsubscript{25\%}, for example in Llama-3-8B, represents replace first 8 layers to D-ReFT and remain the left 24 layers as ReFT intervention.}
        \label{tab:math_result}
    \adjustbox{max width=\textwidth}{
    \begin{tabular}{llrrrrrrrrl}
        \toprule
        \multirow{2}{*}{\textbf{Model}} & \multirow{2}{*}{\textbf{PEFT}} & \multirow{2}{*}{\textbf{Params} (\%)} & \multicolumn{8}{c}{\textbf{Accuracy} ($\uparrow$)} \\
        \cmidrule{4-11}
         \rowcolor{gray!15}
        & & & \textbf{MultiArith}& \textbf{GSM8K}& \textbf{SVAMP}& \textbf{MAWPS} & \textbf{AddSub} & \textbf{AQuA} & \textbf{SingleEq} & \textbf{Avg.} \\
        \midrule
        \multirow{4}{*}{LLaMA-7B} 
        & LoRA & 0.826\% & 88.7 & 25.4 & 46.9 & 74.2 & 82.7 &22.7 &78.1 & 59.8\\
        & RED & 0.039\% & 89.1 & 23.2 & 45.2 & 73.1 & 83.1 &21.6 &\textbf{79.3} & 58.9\\
        & ReFT & 0.031\% & 89.3 & 25.0 & 44.7 & 76.5 & \textbf{83.3} &23.2 &77.6 & 59.9\\ \cmidrule{2-11}
         \rowcolor{gray!15}
        & \textbf{D-ReFT}\textsubscript{25\%} (Ours) & 0.046\% & \textbf{89.5} & \textbf{26.0} & \textbf{47.7}& \textbf{77.2} & 83.0 &\textbf{26.4} &78.9 & \textbf{61.2}\\
        \midrule
        \multirow{4}{*}{LLaMA-2-7B} 
        & LoRA & 0.670\% & 88.4 & \textbf{30.3} & 48.4 & 77.4 & 83.2 & 25.6 & 80.2 & 61.9 \\
        & RED & 0.031\% & 90.5 & 29.2 & \textbf{51.6} & 75.9 & 83.2 & 26.2 & 81.6 & 62.0 \\
        
          & ReFT & 0.025\% & 90.2 & 29.6 & 49.7 & 77.7 & 84.8 & 24.0 & 81.9  & 62.3\\
 \cmidrule{2-11}
  \rowcolor{gray!15}
        & \textbf{D-ReFT}\textsubscript{25\%} (Ours) & 0.037\% & \textbf{91.7} & 30.2 & 51.0 & \textbf{78.2} & \textbf{85.6} & \textbf{26.7} & \textbf{82.5} & \textbf{63.7} \\
            \midrule
        \multirow{4}{*}{LLaMA-3-8B} 
        & LoRA & 0.670\% & 98.0 & 61.3 & 71.7& 89.0 & 92.7 &30.3 & 91.5 & 76.3 \\
        & RED & 0.031\% & 97.8 & 58.2 & 72.0& 88.9 & 92.9 &\textbf{31.0} & 92.7 & 75.9 \\
          & ReFT & 0.025\% & \textbf{98.5} & 60.0 & 72.3& 89.1 & 93.2 &30.0 & 93.1 & 76.6\\
 \cmidrule{2-11}
 \rowcolor{gray!15}
        & \textbf{D-ReFT}\textsubscript{25\%} (Ours) & 0.037\% & 97.4 & \textbf{61.7} & \textbf{73.6} & \textbf{91.6} & \textbf{93.4} & 30.3 & \textbf{94.1} & \textbf{77.4}\\    
        \bottomrule\\
    \end{tabular}
    }
  \vspace{-0.8cm}  
\end{table*}
\paragraph{Interpretation.} This pheonmenon could be understood as early layers in language models has the rich relevant information of the input $X$. By applying distribution-level interventions to these layers, the model retains richer, more flexible representations of uncertainty, avoiding premature over-commitment to specific features. This stochasticity allows the model to propagate diverse hypotheses downstream, which later layers—specializing in high-level reasoning and task-specific logic—can refine using deterministic, pointwise interventions.

\subsection{Arithmetic Reasoning}
We evaluate the performance of our proposed distribution-level intervention method based on ReFT against existing PEFT and intervention methods (LoRA, RED, and ReFT) on arithmetic reasoning datasets using LLaMA-1 7B, LLaMA-2 7B, and LLaMA-3 8B (see Table~\ref{tab:math_result}). Our method consistently outperforms baselines while maintaining parameter efficiency. Notably, D-ReFT$_{25\%}$ achieves the highest average accuracy across all model sizes and excels in key tasks such as GSM8K and SingleEq. These results demonstrate its effectiveness in improving arithmetic reasoning with minimal parameter overhead, making it a promising lightweight replacement for existing ReFT methods. 

\subsection{Influence of Params (\%) in Intervention}
A potential concern is that D-interventions introduce extra parameters for variance calculation, potentially explaining the performance gains. We conduct an ablation study on the \textit{subspace rank} ($r$) to control how many parameters are used in the experiment. We set the rank to 8, 16, 32, 64, and 128, increasing the parameter count by 2$\times$ in each setting.

\begin{figure}[h]
\begin{center}
        \centering 
        \includegraphics[width=\linewidth]{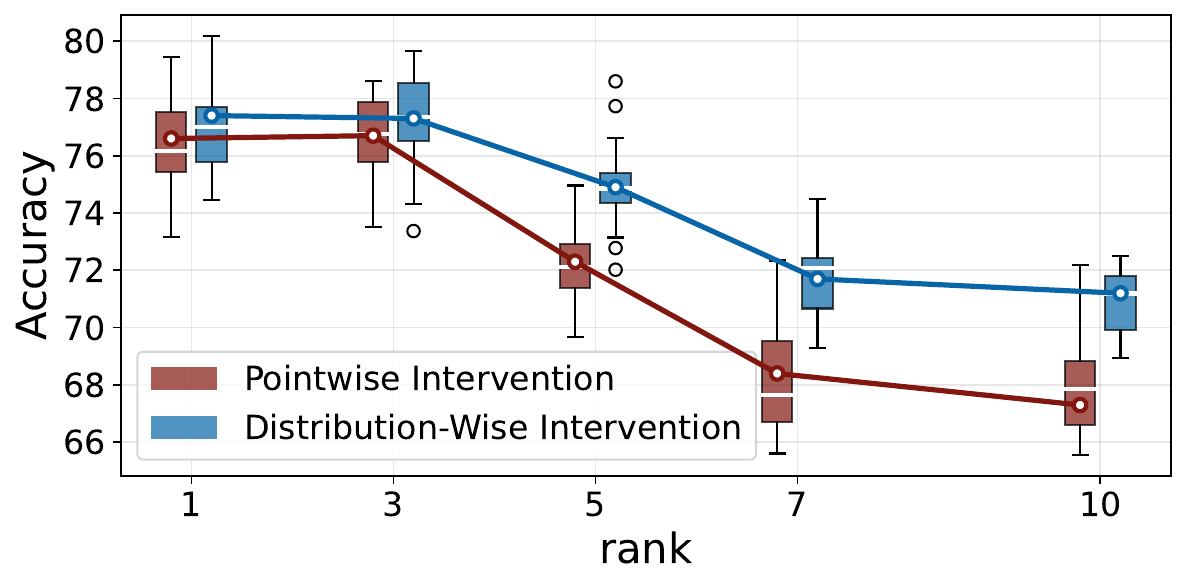}
        \vspace{-0.8cm}
        \caption{Accuracy with different choices of rank in Llama-3-8B for arithmetic reasoning tasks. }
        \label{fig:rank-acc}
\end{center}
\vspace{-0.4cm}
\end{figure}

Analyzing the results of low-rank settings (Figure~\ref{fig:rank-acc}), we first find that ReFT's performance does not improve with higher ranks (more parameters); instead, all methods suffer from a decrease in performance. This suggests that parameter quantity alone fails to drive gains. Both ReFT and D-ReFT peak at ranks 8 and 16, underscoring the efficacy of targeted interventions in lower-dimensional subspaces over sheer parameter growth.

\subsection{Robustness Evaluation}
During our preliminary studies, we tried both synonym replacement (using WordNet~\citep{miller-1994-wordnet}) and paraphrase generation (using back translation). However, our empirical analysis revealed that these semantically-preserving transformations produced insufficient perturbation magnitude to effectively distinguish between intervention methodologies. Therefore, we implemented a more challenging setting to evaluate the robustness of ReFT and D-ReFT variants on arithmetic tasks by randomly deleting $N$ non-arithmetic words from the benchmark and observing the influence on accuracy.

\begin{figure}[h]
\begin{center}
        \centering 
        \includegraphics[width=\linewidth]{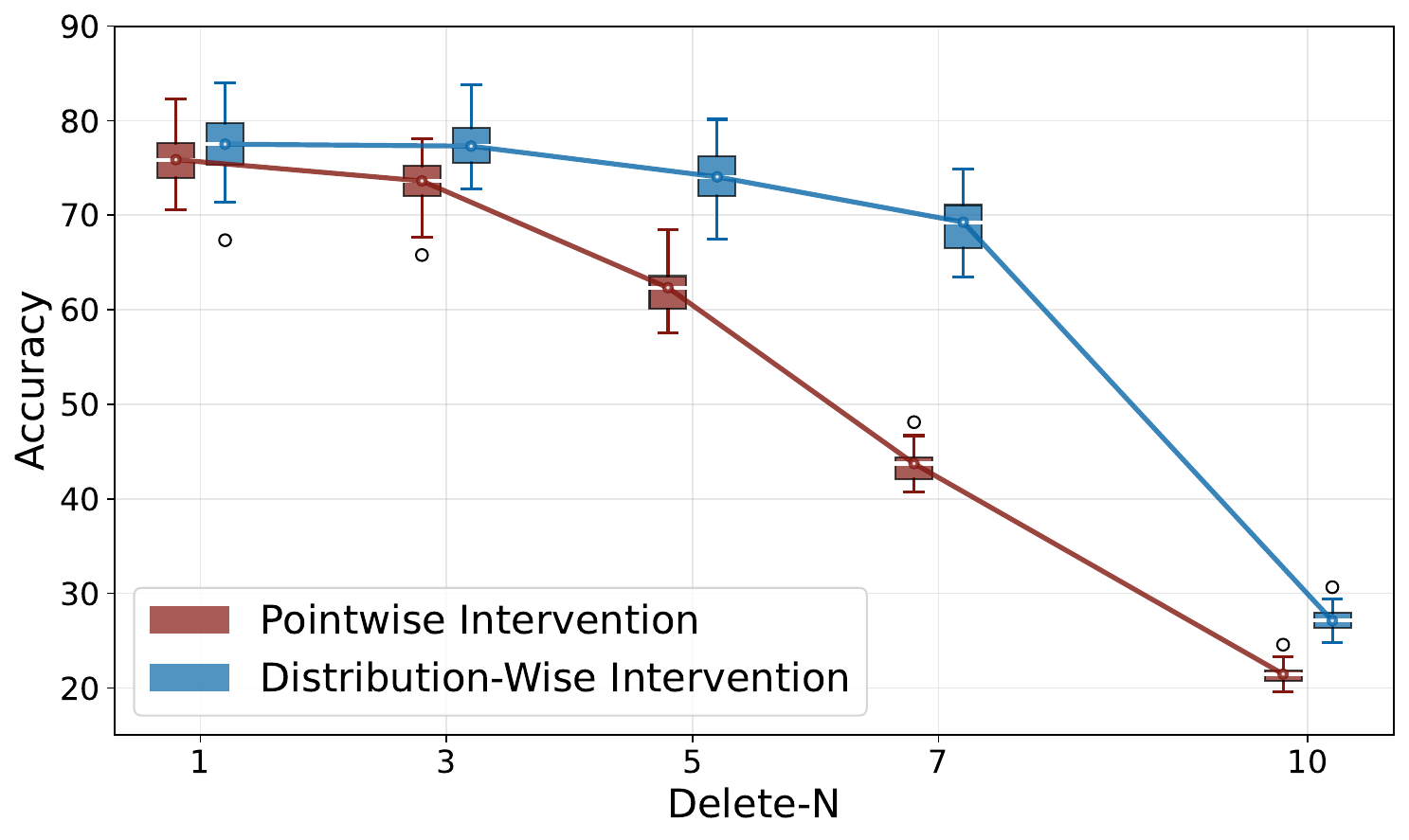}
        \vspace{-0.4cm}
        \caption{Accuracy with randomly deleting $N$ words for testset in Llama-3-8B for arithmetic reasoning tasks.}
        \label{fig:robustness-acc}
\end{center}
\vspace{-0.3cm}
\end{figure}

As D-ReFT learns a distribution over ReFT intervention, we find it exhibits greater robustness to these perturbations (see Figure~\ref{fig:robustness-acc}). When fewer than 8 words are deleted, the accuracy for D-ReFT remains stable, while its pointwise variant drops by approximately 30\%. Although deleting more than 10 words leads to a significant accuracy decline for both methods, the distribution-level variants still demonstrate notable robustness against adversarial attacks. This suggests that distribution-level interventions additionally provide advantages in terms of robustness.

\section{Test-Time Stochasticity: Controlled Temperature Scaling}
\label{sec:temperature_scaling}

While our distribution-wise interventions demonstrate significant improvements during training, a critical question remains: how should the learned stochastic distributions be utilized during inference? We then investigate \textit{controlled stochasticity} through temperature scaling, which allows fine-grained control over the degree of randomness at test time.

We introduce a temperature parameter $\tau$ that scales the learned variance during inference:
\begin{equation}
\vspace{-0.2cm}
\hat{Z} = \boldsymbol{\mu} + (\tau \cdot \boldsymbol{\sigma}) \odot \boldsymbol{\epsilon}, \quad \epsilon \sim \mathcal{N}(0, I)
\end{equation}
where $\tau \geq 0$ controls the stochasticity level:
\begin{align}
\vspace{-0.2cm}
\tau = 0 &: \text{Deterministic inference} \quad (\hat{Z} = \mu) \\
\tau = 1 &: \text{Training-time stochasticity} \\
\tau > 1 &: \text{Increased exploration} \\
0 < \tau < 1 &: \text{Reduced stochasticity}
\vspace{-0.2cm}
\end{align}
We set the $\tau$ to $0, 0.5, 1, 2$ as the ablation study for this section. Specifically we introduce the experiment of instruction-tuning to observe the task difference w.r.t different setting. We use Alpaca-Eval v1.0~\cite{alpaca_eval} for instruction tuning. By default, version 1.0 calculates the win rate against text-davinci-003, with GPT-4 serving as the judge. The prompt template is provided by Alpaca-Eval, and all models in the Alpaca-Eval benchmark use this template for evaluation. For training, we use UltraFeedback~\citep{cui2024ultrafeedbackboostinglanguagemodels}, a high-quality instruction-tuning dataset that covers various aspects like general IT knowledge, truthfulness, honesty, and helpfulness to assess model performance. This setup aligns with the previous work on RED and ReFT.

We adopt the recommended hyperparameter settings from the paper for baseline methods like ReFT. For D-ReFT, we directly applied the params used in the math arithmetic learning datasets. All results are reported over three runs.

\paragraph{Results.}
Table~\ref{tab:temperature_gain_tasks} reveals distinct patterns in how temperature scaling affects different task categories. Lower temperature values ($\tau < 1$) consistently improve performance on commonsense and arithmetic reasoning tasks, with deterministic inference ($\tau = 0$) achieving the largest gains of $+0.7$ and $+0.6$ points respectively. 
\begin{table}[h]
    \centering
    \vspace{-0.65cm}
    \caption{Performance of D-ReFT with Different Temperature Values Across Task Types.}
    \vspace{+0.2cm}
    \adjustbox{max width=0.48\textwidth}{
    \begin{tabular}{lccc}
        \toprule
        \textbf{Method} & \textbf{Commonsense} & \textbf{Arithmetic} & \textbf{Instruction Following} \\
        \midrule
        \rowcolor{gray!15}
        D-ReFT (Baseline) & 88.6 & 76.6 & 82.4 \\
        ~~~+~$\mathit{\tau = 0}$ & 89.3 & 79.2 & 80.1 \\
        ~~~+~$\mathit{\tau = 0.5}$ & 88.7 & 78.4 & 81.8 \\
        ~~~+~$\mathit{\tau = 1}$ & 86.9 & 76.8 & 83.7 \\
        ~~~+~$\mathit{\tau = 2}$ & 85.1 & 75.2 & 85.9 \\
        \bottomrule
    \end{tabular}}
    \label{tab:temperature_gain_tasks}
\vspace{-0.2cm}
\end{table}
Conversely, these same low-temperature settings degrade instruction following performance, suggesting that deterministic processing may be too rigid for the diverse response patterns required in instruction-based tasks. Higher temperature values ($\tau \geq 1$) reverse this trend, with $\tau = 2$ providing substantial improvements in instruction following while simultaneously degrading performance on reasoning tasks that benefit from more focused, consistent processing.

\section{Conclusion}
In this work, we introduced a distribution-wise intervention framework that extends traditional pointwise intervention methods for modifying language model representations. By replacing deterministic nodes with stochastic ones, our approach enables more robust and fine-grained control in the latent space. Through comprehensive evaluations across multiple commonsense and arithmetic reasoning benchmarks, we demonstrated that distribution-level interventions significantly improve controllability and robustness, particularly in early layers of the model. Our results suggest that incorporating distribution-aware modifications into model training could be a promising direction for improving interpretability and steering model behavior with greater precision and enable finer control.

\section*{Acknowledgment}
We thank the anonymous reviewers for their valuable comments. We thank Zhengxuan Wu for the kind assistance with the ReFT codebase, which is convenient for intervention research. We also thank the members of the Chili Lab for their valuable suggestions for the work and writing. 
\section*{Impact Statement}
This paper presents work whose goal is to advance the field of 
Machine Learning. There are many potential societal consequences 
of our work, none which we feel must be specifically highlighted here.

\bibliography{custom}
\bibliographystyle{icml2025}

\newpage
\appendix
\onecolumn
\section{Data Processing Inequality}
\label{appendix:dpi}
The Data Processing Inequality (DPI) is a fundamental concept in information theory that describes the loss of information through a processing pipeline. In deep neural network, the mutual information satisfies:
\begin{equation}
 I(Y; X) \geq  I(Y; Z)
\end{equation}
where \(I(\cdot; \cdot)\) denotes the mutual information between two random variables.

Intuitively, this inequality implies that processing data, represented by the transition from \(Z\) to \(Y\), cannot increase the information about the original variable \(X\). In other words, no transformation or operation on \(Z\) can recover information about \(X\) that has already been lost.
In the context of LMs, this concept is particularly relevant. The hidden representations \(Z^{(l)}\) at layer \(l\) encode information about the input sequence \(X\). As these representations are transformed layer by layer through attention mechanisms and feed-forward networks, the DPI suggests that the mutual information between the input \( X \) and the final output \( Y \) is non-increasing as we move deeper into the network:
\begin{equation}
\displaystyle I(Y; X) \geq I(X; Z^{(L)}) \geq \cdots \geq I(X; Z^{(1)}) \geq I(Y;\hat{Y})
\end{equation}

\textit{This highlights a trade-off in deep architectures}: while deeper layers may refine representations for specific tasks, they cannot recover information lost in earlier layers. 

By the chain rule of mutual information, minimizing \(\mathcal{L}_{\text{CE}}\) maximizes \(I(Y; \hat{Y})\) while implicitly encouraging the learned representations \(Z^{(L)}\) to retain sufficient task-relevant information about \(Y\). 
\section{Hyperparameter Configuration}
\label{appendix:hyperparameter}
In this section we will discuss all the hyperparameter setting for previous method.

\begin{tcolorbox}[colback=gray!10, colframe=gray!50, title={Hyperparameter Configuration}]
We investigate the following hyperparameters for our experiments:
\begin{itemize}
    \item \textbf{Intervention Layer} ($l$): The specific layer in the model where the intervention is applied. This is chosen based on the architecture and the desired impact on the model's behavior.
    \item \textbf{Noise Scale} ($\epsilon$): The magnitude of noise added during the intervention. This controls the level of perturbation introduced to the model's activations.
    \item \textbf{Subspace Rank} ($r$): The rank of the subspace used for the intervention. This determines the dimensionality of the subspace in which the intervention operates.
    \item \textbf{Intervention Position} ($p$): The position within the layer where the intervention is applied (e.g., before or after a specific operation like activation or normalization).
    \item \textbf{Batch Size} ($bs$): The number of samples processed in each batch during training. This affects the stability and speed of the training process.
    \item \textbf{Training Epochs} ($e$): The total number of times the model is trained over the entire dataset. This influences the convergence and generalization of the model.
    \item \textbf{Learning Rate} ($lr$): The step size at which the model's parameters are updated during training. This controls the speed and stability of the learning process.
\end{itemize}
\end{tcolorbox}

\subsection{ReFT}
\begin{tcolorbox}[colback=gray!10, colframe=gray!50, title={ReFT Hyperparameter Configuration}]
We investigate the following hyperparameters for our experiments:
\begin{itemize}
    \item \textbf{Learning Rate} ($lr$): $9e-4$. 
    \item \textbf{Subspace Rank} ($r$): $8$ or $16$ works best. Higher rank like $256$ does not introduce the boost. 
    \item \textbf{Intervention Position} ($p$): $f7+l7$: first seven token and last seven token.
    \item \textbf{Batch Size} ($bs$): $8$. We also enable gradient checkpoint with accumulated steps = $4$. We are unable to ablate this params due to memory constraints.
    \item \textbf{Training Epochs} ($e$): $12$ works best. In our experiments decrease training epochs to $9$ lead to performance drop.
      \item \textbf{Intervention Layer} ($l$): Depends on the experiment setting, but we find early layers work best.
\end{itemize}
\end{tcolorbox}
\subsection{D-ReFT}
\begin{tcolorbox}[colback=gray!10, colframe=gray!50, title={ReFT Hyperparameter Configuration}]
We investigate the following hyperparameters for our experiments:
\begin{itemize}
    \item \textbf{Learning Rate} ($lr$): $1e-3$ or $3e-3$. 

    \item \textbf{Subspace Rank} ($r$): $8$ or $16$ works best. Higher rank like $256$ does not introduce the boost. 
    \item \textbf{Intervention Position} ($p$): $f7+l7$: first seven token and last seven token.
    \item \textbf{Batch Size} ($bs$): $8$. We also enable gradient checkpoint with accumulated steps = $4$. We are unable to ablate this params due to memory constraints.
    \item \textbf{Training Epochs} ($e$): $9$ works best, which showcases that D-ReFT also have better convergence property.
     \item \textbf{Intervention Layer} ($l$): Depends on the experiment setting, but we find early layers work best.
\end{itemize}
\end{tcolorbox}
\subsection{LoRA}

\begin{tcolorbox}[colback=gray!10, colframe=gray!50, title={ReFT Hyperparameter Configuration}]
We investigate the following hyperparameters for our experiments:
\begin{itemize}
    \item \textbf{Learning Rate} ($lr$): $4e-4$. 
          \item \textbf{Alpha} ($\alpha$): $16$. 

    \item \textbf{Subspace Rank} ($r$): $16$ works best.  
    \item \textbf{Intervention Position} ($p$): $all$: all token positions are intervented.
    \item \textbf{Batch Size} ($bs$): $8$. We also enable gradient checkpoint with accumulated steps = $4$. We are unable to ablate this params due to memory constraints.
    \item \textbf{Training Epochs} ($e$): $6$ works best.
     \item \textbf{Intervention Layer} ($l$): Depends on the experiment setting, but we find early layers work best.
\end{itemize}
\end{tcolorbox}
\subsection{RED.}

\begin{tcolorbox}[colback=gray!10, colframe=gray!50, title={RED Hyperparameter Configuration}]
We investigate the following hyperparameters for our experiments:
\begin{itemize}
    \item \textbf{Learning Rate} ($lr$): $7e-4$. 

    \item \textbf{Intervention Position} ($p$): $all$: all token positions are intervented.
    \item \textbf{Batch Size} ($bs$): $8$. We also enable gradient checkpoint with accumulated steps = $4$. We are unable to ablate this params due to memory constraints.
    \item \textbf{Training Epochs} ($e$): $9$ works best.
     \item \textbf{Intervention Layer} ($l$): Depends on the experiment setting, but we find early layers work best.
\end{itemize}
\end{tcolorbox}

\subsection{SwiGLU.}

\begin{tcolorbox}[colback=gray!10, colframe=gray!50, title={SwiGLU Hyperparameter Configuration}]
We investigate the following hyperparameters for our experiments:
\begin{itemize}
    \item \textbf{Learning Rate} ($lr$): $6e-4$. 

    \item \textbf{Intervention Position} ($p$): $all$: all token positions are intervented.
    \item \textbf{Batch Size} ($bs$): $8$. We also enable gradient checkpoint with accumulated steps = $4$. We are unable to ablate this params due to memory constraints.
    \item \textbf{Training Epochs} ($e$): $9$ works best.
     \item \textbf{Intervention Layer} ($l$): Depends on the experiment setting, but we find early layers work best.
\end{itemize}
\end{tcolorbox}

\subsection{MLP.}

\begin{tcolorbox}[colback=gray!10, colframe=gray!50, title={MLP Hyperparameter Configuration}]
We investigate the following hyperparameters for our experiments:
\begin{itemize}
    \item \textbf{Learning Rate} ($lr$): $5e-4$. 

    \item \textbf{Intervention Position} ($p$): $all$: all token positions are intervented.
    \item \textbf{Batch Size} ($bs$): $8$. We also enable gradient checkpoint with accumulated steps = $4$. We are unable to ablate this params due to memory constraints.
    \item \textbf{Training Epochs} ($e$): $9$ works best.
     \item \textbf{Intervention Layer} ($l$): Depends on the experiment setting, but we find early layers work best.
\end{itemize}
\end{tcolorbox}

\section{Datasets.}
\label{appendix:datasets}
In this section we will introduce the benchmarks we are use in this paper. Generally, we follow the setting of ~\citet{hu-etal-2023-llm,wu2024reftrepresentationfinetuninglanguage}, using eight commonsense benchmarks and seven arithmetic benchmarks to evaluate. 
\subsection{Commonsense Reasoning}
\paragraph{BoolQ.}
The BoolQ~\cite{clark2019boolqexploringsurprisingdifficulty} dataset is a collection of natural language questions and corresponding passages designed for the task of binary question answering. It contains over 15,000 examples where each question can be answered with a simple "yes" or "no" based on the information provided in the accompanying passage. The dataset is derived from real user queries and web pages, making it a valuable resource for training and evaluating models on understanding context and reasoning over text. BoolQ is widely used in NLP research to benchmark the performance of models in tasks requiring comprehension, inference, and binary classification. Its challenging nature stems from the need for models to grasp nuanced relationships between questions and passages, making it a key dataset for advancing question-answering systems.

\begin{tcolorbox}[colback=gray!10, colframe=gray!50, title={An example of data from the BoolQ dataset}]
\textbf{Instructions:} Please answer the following question with true or false, question: does ethanol take more energy make that produces?\\\textbf{Answer format:} true/false.
\end{tcolorbox}

\paragraph{PIQA.}
The PIQA~\cite{bisk2019piqareasoningphysicalcommonsense}  dataset is a benchmark designed to evaluate a model's understanding of physical commonsense reasoning in everyday scenarios. It consists of questions that require reasoning about how objects interact, are used, or are manipulated in the physical world. Each question presents two possible solutions to a practical problem, and the task is to select the most appropriate one based on real-world physics and intuition. PIQA challenges models to go beyond textual knowledge and incorporate an understanding of physical properties, causality, and affordances, making it a valuable resource for advancing AI systems in tasks that require grounded, real-world reasoning.
\begin{tcolorbox}[colback=gray!10, colframe=gray!50, title={An example of data from the PIQA dataset}]
\textbf{Instructions:} Please choose the correct solution to the question: How do I ready a guinea pig cage for it's new occupants?\\\textbf{Solution1}: Provide the guinea pig with a cage full of a few inches of bedding made of ripped paper strips, you will also need to supply it with a water bottle and a food dish.\\\textbf{Solution2:} Provide the guinea pig with a cage full of a few inches of bedding made of ripped jeans material, you will also need to supply it with a water bottle and a food dish.\\\textbf{Answer format:} solution1 or solution2
\end{tcolorbox}

\paragraph{HellaSwag.}
The HellaSwag~\cite{zellers2019hellaswagmachinereallyfinish} dataset is a benchmark designed to evaluate the commonsense reasoning capabilities of natural language understanding models. Introduced in 2019, it consists of multiple-choice questions that require models to predict the most plausible continuation of a given scenario, drawing on everyday knowledge and contextual understanding. Unlike many other datasets, HellaSwag emphasizes real-world situations and nuanced reasoning, making it particularly challenging for state-of-the-art models. The dataset was created to address the limitations of previous benchmarks, which often relied on superficial patterns or biases in the data. By focusing on scenarios that require deeper comprehension and inference, HellaSwag has become a valuable tool for advancing research in artificial intelligence and improving the robustness of language models.

\begin{tcolorbox}[colback=gray!10, colframe=gray!50, title={An example of data from the HellaSwag dataset}]
\textbf{Instructions:} Please choose the correct ending to complete the given sentence: Roof shingle removal: A man is sitting on a roof. he\\\textbf{Ending1:} is using wrap to wrap a pair of skis. \\\textbf{Ending2:} is ripping level tiles off. \\\textbf{Ending3:} is holding a rubik's cube. \\ \textbf{Ending4}: starts pulling up roofing on a roof.\\\textbf{Answer format}: ending1/ending2/ending3/ending4.
\end{tcolorbox}

\paragraph{WinoGrande.}
The WinoGrande~\cite{sakaguchi2019winograndeadversarialwinogradschema} dataset is a large-scale collection of natural language inference problems designed to evaluate the reasoning capabilities of artificial intelligence systems, particularly in the context of commonsense reasoning. Introduced as a more challenging successor to the Winograd Schema Challenge, WinoGrande contains over 44,000 carefully crafted pronoun resolution problems that require understanding context, world knowledge, and subtle linguistic cues. Each problem presents a short passage with an ambiguous pronoun, and the task is to determine the correct referent from two possible options. To address biases and ensure robustness, the dataset was created using a crowdsourcing approach followed by a systematic adversarial filtering process. WinoGrande has become a benchmark for testing the limits of machine learning models in handling complex reasoning tasks, pushing the boundaries of AI systems toward more human-like comprehension and decision-making.

\begin{tcolorbox}[colback=gray!10, colframe=gray!50, title={An example of data from the WinoGrande dataset}]
\textbf{Instructions:} Please choose the correct answer to fill in the blank to complete the given sentence: Sarah was a much better surgeon than Maria so always got the easier cases.\\ \textbf{Option1}: Sarah \\ \textbf{Option2}: Maria \\\textbf{Answer format}: option1/option2
\end{tcolorbox}

\paragraph{ARC-e.}
The ARC-e (AI2 Reasoning Challenge - Easy)~\cite{clark2018thinksolvedquestionanswering} dataset is a collection of elementary-level science questions designed to evaluate the reasoning and comprehension capabilities of artificial intelligence systems. Developed by the Allen Institute for AI, ARC-e focuses on multiple-choice questions that require a fundamental understanding of scientific concepts, making it an accessible yet challenging benchmark for AI models. Unlike its more advanced counterpart, ARC (AI2 Reasoning Challenge), ARC-e is tailored to assess basic knowledge and straightforward reasoning, often drawing from topics taught in early education. By providing a simplified yet diverse set of questions, ARC-e serves as a valuable tool for testing the foundational abilities of AI systems in processing and answering science-related queries, paving the way for more complex reasoning tasks.

\begin{tcolorbox}[colback=gray!10, colframe=gray!50, title={An example of data from the ARC-e dataset}]
\textbf{Instructions:} Please choose the correct answer to the question: Which statement best explains why photosynthesis is the foundation of most food webs?\\ \textbf{Answer1}: Sunlight is the source of energy for nearly all ecosystems. \\ \textbf{Answer2}: Most ecosystems are found on land instead of in water. \\ \textbf{Answer3}: Carbon dioxide is more available than other gases. \\ \textbf{Answer4}: The producers in all ecosystems are plants.\\ \textbf{Answer format}: answer1/answer2/answer3/answer4
\end{tcolorbox}

\paragraph{ARC-c.}
The ARC-c~\cite{clark2018thinksolvedquestionanswering} dataset, part of the AI2 Reasoning Challenge (ARC), is a comprehensive collection of science questions designed to evaluate the reasoning and comprehension capabilities of artificial intelligence systems. Comprising multiple-choice questions from various grade levels, the dataset emphasizes complex reasoning, requiring models to go beyond simple retrieval and engage in deeper understanding and inference. The questions are drawn from diverse scientific domains, including biology, chemistry, physics, and earth science, making it a robust benchmark for assessing the generalization and problem-solving skills of AI. By focusing on challenging, curriculum-aligned content, the ARC-c dataset serves as a critical tool for advancing the development of AI systems capable of nuanced and context-aware reasoning.
\begin{tcolorbox}[colback=gray!10, colframe=gray!50, title={An example of data from the ARC-c dataset}]
\textbf{Instructions:} Please choose the correct answer to the question: A group of engineers wanted to know how different building designs would respond during an earthquake. They made several models of buildings and tested each for its ability to withstand earthquake conditions. Which will most likely result from testing different building designs?\\\textbf{Answer1}: buildings will be built faster \\ \textbf{Answer2}: buildings will be made safer \\ \textbf{Answer3}: building designs will look nicer \\ \textbf{Answer4:} building materials will be cheaper\\ \textbf{Answer format}: answer1/answer2/answer3/answer4
\end{tcolorbox}

\paragraph{OBQA.}
The Open Book Question Answering (OBQA)~\cite{mihaylov2018suitarmorconductelectricity} dataset is a benchmark designed to evaluate the ability of machine learning models to answer science-based questions by combining open-book fact retrieval with reasoning skills. Unlike traditional QA datasets, OBQA requires systems to not only retrieve relevant information from a provided knowledge source but also apply logical reasoning to infer the correct answer. The dataset consists of multiple-choice questions that span a wide range of scientific topics, challenging models to demonstrate both comprehension and analytical capabilities. By emphasizing the integration of external knowledge and reasoning, OBQA serves as a valuable tool for advancing research in question answering, knowledge representation, and artificial intelligence.

\begin{tcolorbox}[colback=gray!10, colframe=gray!50, title={An example of data from the OBQA dataset}]
\textbf{Instructions:} Please choose the correct answer to the question: A person wants to start saving money so that they can afford a nice vacation at the end of the year. After looking over their budget and expenses, they decide the best way to save money is to\\ \textbf{Answer1}: make more phone calls \\ \textbf{Answer2}: quit eating lunch out \\ \textbf{Answer3}: buy less with monopoly money \\ \textbf{Answer4}: have lunch with friends \\ \textbf{Answer format}: answer1/answer2/answer3/answer4
\end{tcolorbox}

\subsection{Arithmetic Reasoning}

\paragraph{AQuA.}
The AQuA (Algebraic Question Answering)~\cite{ling2017programinductionrationalegeneration} dataset is a specialized resource designed to advance research in the field of mathematical reasoning and natural language processing. It consists of a collection of algebra word problems, each paired with multiple-choice answers and detailed step-by-step solutions. The dataset is particularly valuable for training and evaluating models that aim to understand and solve complex mathematical problems expressed in natural language. By providing both the problem statements and their corresponding solutions, AQuA enables researchers to develop and test algorithms that can not only generate correct answers but also explain the reasoning process behind them. This makes AQuA an essential tool for exploring the intersection of mathematics and artificial intelligence, fostering advancements in automated problem-solving and educational technologies.

\begin{tcolorbox}[colback=gray!10, colframe=gray!50, title={An example of data from the AQuA dataset}]
\paragraph{Instructions:} A rectangular piece of 150 sq m has a length which is 1m more than the 4 times the breadth. What is the perimeter of the piece? Answer Choices: (A) 60 m (B) 61 m (C) 62 m (D) 63 m (E) 64 m \\
Let's assume the breadth of the rectangle to be \( x \). According to the problem, the length of the rectangle is \( 1 \) m more than \( 4 \) times the breadth. Therefore, the length of the rectangle is \( 4x + 1 \).

We know that the area of the rectangle is \( 150 \) sq m. Therefore, we can write:
\[
\text{Area} = \text{Length} \times \text{Breadth}
\]
\[
150 = (4x + 1) \times x
\]
Simplifying this equation, we get:
\[
150 = 4x^2 + x
\]
\[
150 - x = 4x^2
\]
\[
x^2 - \frac{1}{4}x - \frac{15}{4} = 0
\]
Solving for \( x \), we get:
\[
x = 1.732 \quad \text{or} \quad x = -0.732
\]
Since the breadth cannot be negative, we can ignore the negative value of \( x \). Therefore, the breadth of the rectangle is \( 1.732 \) m.

Now, we can find the length of the rectangle:
\[
\text{Length} = 4x + 1 = 4(1.732) + 1 = 7.148 \, \text{m}
\]
The perimeter of the rectangle is the sum of the length of all four sides. Therefore, the perimeter is:
\[
\text{Perimeter} = 2(\text{Length} + \text{Breadth}) = 2(7.148 + 1.732) = 16.928 \, \text{m}
\]
Rounding off to the nearest integer, we get the answer as \( 17 \). Therefore, the answer is \(\boxed{(E) \, 64 \, \text{m}}\).
\end{tcolorbox}

\paragraph{AddSub.}
The AddSub~\cite{hosseini-etal-2014-learning} dataset is a widely-used benchmark in NLP designed to evaluate the ability of models to solve arithmetic word problems. It consists of pairs of questions and answers, where each question is a textual description of a mathematical problem involving addition or subtraction, and the corresponding answer is the numerical result. This dataset challenges models to not only understand the linguistic nuances of the problem but also to perform the necessary calculations accurately. By focusing on basic arithmetic operations, AddSub serves as a fundamental testbed for assessing the reasoning and comprehension capabilities of NLP systems, making it a valuable resource for research in machine learning and artificial intelligence.

\begin{tcolorbox}[colback=gray!10, colframe=gray!50, title={An example of data from the AddSub dataset}]
\paragraph{Instructions:} There are 7 crayons in the drawer . Mary took 3 crayons out of the drawer. How many crayons are there now? \\
Step 1: Start with the total number of crayons in the drawer: 7\\Step 2: Subtract the number of crayons Mary took out: 3\\Step 3: Perform the subtraction: 7 - 3 = 4\\\textbf{Answer}: There are now 4 crayons in the drawer.
\end{tcolorbox}

\paragraph{GSM8K.}
The GSM8K~\cite{cobbe2021trainingverifierssolvemath} dataset is a collection of 8.5K high-quality, linguistically diverse grade school math word problems designed to benchmark the problem-solving abilities of AI systems. Each problem in the dataset requires multiple reasoning steps to arrive at the correct solution, making it a valuable resource for evaluating the mathematical and logical reasoning capabilities of language models. The problems are crafted to reflect real-world scenarios, ensuring that they are both challenging and accessible to students at the elementary level. By providing a diverse set of problems that span various mathematical concepts, GSM8K serves as a robust testbed for advancing the development of AI systems capable of understanding and solving complex, multi-step problems.

\begin{tcolorbox}[colback=gray!10, colframe=gray!50, title={An example of data from the GSM8K dataset}]
\paragraph{Instructions:} Mr Boarden is remodeling his bathroom. For every square foot, he needs 24 mosaic tiles. How many mosaic tiles would Mr Boarden need to cover two thirds of his 36 sq ft bathroom? \\
Step 1: Find the total area of the bathroom: 36 sq ft\\Step 2: Calculate the area to be covered by the mosaic tiles\\36 sq ft x 2/3 = 24 sq ft\\Step 3: Calculate the number of mosaic tiles needed\\24 sq ft / 1 sq ft per 24 tiles = 1 tile\\Therefore, Mr Boarden would need 1 mosaic tile to cover two thirds of his 36 sq ft bathroom. \\The answer in Arabic numerals is 1.
\end{tcolorbox}

\paragraph{MAWPS.}
The MAWPS (Math Word Problem Solving)~\cite{koncel-kedziorski-etal-2016-mawps}, is a comprehensive collection of mathematical word problems designed to facilitate research in NLP. This dataset encompasses a wide range of problem types, including arithmetic, algebra, and geometry, presented in natural language form, making it a valuable resource for developing and evaluating algorithms that aim to solve math problems through automated reasoning and language understanding. MAWPS is particularly useful for training and testing models that require both linguistic comprehension and mathematical problem-solving skills, bridging the gap between textual data and quantitative reasoning. By providing a diverse set of problems with varying levels of complexity, MAWPS serves as a benchmark for advancing the capabilities of AI systems in educational applications and beyond.

\begin{tcolorbox}[colback=gray!10, colframe=gray!50, title={An example of data from the MAWPS dataset}]
\paragraph{Instructions:} Nicole found an orange caterpillar and a green caterpillar in her backyard . The green caterpillar was 3 inches long and the orange caterpillar was $1.1666666666666667$ inches long . How much longer was the green caterpillar than the orange caterpillar ? \\
To find the difference in length between the green and orange caterpillars, we need to subtract the length of the orange caterpillar from the length of the green caterpillar.\\Length of green caterpillar - Length of orange caterpillar = Difference in length\\3 inches - 1.1666666666666667 inches = 1.8333333333333333 inches\\Therefore, the green caterpillar was 1.8333333333333333 inches longer than the orange caterpillar. \\The answer in Arabic numerals is 1.8333333333333333
\end{tcolorbox}

\paragraph{MultiArith.}
The MultiArith~\cite{roy2016solvinggeneralarithmeticword} dataset is a benchmark collection designed to evaluate the performance of machine learning models, particularly in solving arithmetic word problems that require multiple steps of reasoning. This dataset consists of a variety of mathematical problems presented in natural language, challenging models to interpret the text, extract relevant numerical information, and perform sequential arithmetic operations to arrive at the correct solution. By encompassing problems that involve addition, subtraction, multiplication, and division, often in combination, MultiArith tests both the linguistic comprehension and mathematical reasoning capabilities of AI systems. It serves as a valuable resource for advancing research in NLP, particularly in bridging the gap between language understanding and quantitative problem-solving.

\begin{tcolorbox}[colback=gray!10, colframe=gray!50, title={An example of data from the MultiArith dataset}]
\paragraph{Instructions:} At the schools book fair Sam bought 13 adventure books and 17 mystery books. If 15 of the books were used, how many new books did he buy? \\
To solve the problem, we need to find the total number of books Sam bought and then subtract the number of used books from that total.\\1. Find the total number of books Sam bought: \\13 adventure books + 17 mystery books = 30 books\\2. Subtract the number of used books from the total: \\30 books - 15 used books = 15 new books\\Therefore, Sam bought 15 new books. \\The answer in Arabic numerals is 15.
\end{tcolorbox}

\paragraph{SingleEq.}
The SingleEq~\cite{koncel-kedziorski-etal-2015-parsing} dataset is a specialized resource designed to support research and development in the field of mathematical problem-solving, particularly focusing on single-variable linear equations. This dataset comprises a collection of problems, each involving the formulation and solution of linear equations with one unknown, making it an invaluable tool for training and evaluating machine learning models, educational software, and automated tutoring systems. By providing a structured and diverse set of equations, the SingleEq dataset enables researchers to explore various computational approaches, from symbolic reasoning to neural network-based methods, aiming to enhance the accuracy and efficiency of automated equation-solving technologies. Its comprehensive nature ensures that models trained on this dataset can handle a wide range of equation types, promoting robustness and generalizability in mathematical problem-solving applications.

\begin{tcolorbox}[colback=gray!10, colframe=gray!50, title={An example of data from the SingleEq dataset}]
\paragraph{Instructions:} Oceanside Bike Rental Shop charges 17 dollars plus 7 dollars an hour for renting a bike. Tom paid 80 dollars to rent a bike. How many hours did he pay to have the bike checked out ? \\
Let's assume that Tom had the bike checked out for \"x\" hours.\\According to the problem, the cost of renting a bike is 17 dollars plus 7 dollars per hour. So, the total cost for x hours can be represented as:\\17 + 7x\\We know that Tom paid 80 dollars to rent the bike, so we can set up an equation:\\17 + 7x = 80\\Subtracting 17 from both sides, we get:\\7x = 63\\Dividing both sides by 7, we get:\\x = 9\\Therefore, Tom paid to have the bike checked out for 9 hours.
\end{tcolorbox}

\paragraph{SVAMP.}
The SVAMP~\cite{patel-etal-2021-nlp} (Simple Variations in Arithmetic Word Problems) dataset is a carefully curated collection of arithmetic word problems designed to evaluate and enhance the problem-solving capabilities of machine learning models. Unlike traditional datasets, SVAMP introduces variations in problem structure, wording, and complexity to test the robustness and generalization of models across diverse scenarios. Each problem is crafted to require a combination of mathematical reasoning and natural language understanding, making it a valuable benchmark for assessing the performance of AI systems in real-world applications. By incorporating a wide range of problem types, SVAMP aims to bridge the gap between simple arithmetic tasks and more complex, context-rich challenges, providing a comprehensive tool for advancing research in mathematical reasoning and natural language processing.
\begin{tcolorbox}[colback=gray!10, colframe=gray!50, title={An example of data from the SVAMP dataset}]
\paragraph{Instructions:} Matthew gave equal numbers of crackers and cakes to his 4 friends. If he had 32 crackers and 98 cakes initially. How many crackers did each person eat? \\
To solve the problem, we need to divide the total number of crackers by the number of friends. \\Step 1: Find the total number of crackers by adding them up. \\32 crackers \\Step 2: Divide the total number of crackers by the number of friends. \\32 crackers $\div$ 4 friends = 8 crackers per person \\Therefore, each person ate 8 crackers. \\Answer: 8
\end{tcolorbox}

\end{document}